\documentclass[sigconf, preprint]{acmart}
\AtBeginDocument{%
  }

\usepackage{xcolor}
\newcommand{\rr}{\color{black}}
\newcommand{\bb}{\color{black}}

\acmSubmissionID{}

\setcopyright{none}             
\renewcommand\footnotetextcopyrightpermission[1]{}
\acmDOI{}                       
\acmISBN{}                      
\acmConference[]{}{}{}          
\acmYear{}

\makeatletter
\let\@authorsaddresses\@empty
\makeatother

\begin{document}

\title{Virtual Scanning for NSCLC Histology: Investigating the Discriminatory Power of Synthetic PET}
\author{Fatih Aksu}
\authornote{Both authors contributed equally to this research.}
\email{fatih.aksu@unicampus.it}
\orcid{0000-0001-5677-2024}
\author{Laura Ciuffetti}
\authornotemark[1]
\email{laura.ciuffetti@alcampus.it}
\affiliation{%
  \institution{Università Campus Bio-Medico di Roma}
  \city{Rome}
  \country{Italy}
}

\author{Francesco Di Feola}
\email{francesco.feola@umu.se}
\affiliation{%
  \institution{Umeå University}
  \city{Umeå}
  \country{Sweden}
}

\author{Filippo Ruffini}
\email{filippo.ruffini@umu.se}
\affiliation{%
  \institution{Umeå University}
  \city{Umeå}
  \country{Sweden}
}

\author{Giulia Romoli}
\email{giulia.romoli@umu.se}
\affiliation{%
  \institution{Umeå University}
  \city{Umeå}
  \country{Sweden}
}

\author{Fabrizia Gelardi}
\email{gelardi.fabrizia@hsr.it}
\affiliation{%
  \institution{Università Vita-Salute San Raffaele}
  \city{Milan}
  \country{Italy}
}

\author{Arturo Chiti}
\email{chiti.arturo@hsr.it}
\affiliation{%
  \institution{Università Vita-Salute San Raffaele}
  \city{Milan}
  \country{Italy}
}

\author{Valerio Guarrasi}
\email{valerio.guarrasi@unicampus.it}
\affiliation{%
  \institution{Università Campus Bio-Medico di Roma}
  \city{Rome}
  \country{Italy}
}

\author{Paolo Soda}
\email{p.soda@unicampus.it}
\affiliation{%
  \institution{Università Campus Bio-Medico di Roma}
  \city{Rome}
  \country{Italy}
}
\email{paolo.soda@umu.se}
\affiliation{%
  \institution{Umeå University}
  \city{Umeå}
  \country{Sweden}
}

\renewcommand{\shortauthors}{Aksu et al.}

\begin{abstract}
Accurate histological differentiation between adenocarcinoma (ADC) and squamous cell carcinoma (SCC) is critical for personalized treatment in non-small cell lung cancer (NSCLC).
While [$^{18}$F]FDG PET/CT is a standard tool for the clinical evaluation of lung cancer, its utility is often limited by high costs and radiation exposure. 
In this paper, we investigate the 
\rr feasibility of "virtual scanning" as a feature-enhancement strategy by 
evaluating whether synthetic PET data can provide complementary feature representations to supplement anatomical CT scans \bb
in histological subtype classification. 

We propose a framework that leverages a 3D Pix2Pix Generative Adversarial Network (GAN), pretrained on the FDG-PET/CT Lesions dataset, to synthesize pseudo-PET volumes from anatomical CT scans. 
These synthetic volumes are integrated with structural CT data within the MINT framework, a multi-stage intermediate fusion architecture.

Our experiments, conducted on a multi-center dataset of 714 subjects, demonstrate that the inclusion of synthetic metabolic features significantly improves classification performance over a CT-only baseline. 
The multimodal approach achieved a statistically significant increase in the Area Under the Curve (AUC) from 0.489 to 0.591 and improved the Geometric Mean (GMean) from 0.305 to 0.524. 
These results suggest that synthetic PET scans provide discriminatory metabolic cues that enable deep learning models to exploit complementary cross-modal information, 
\rr offering a potential feature-enhancement strategy for clinical scenarios where physical PET scans are unavailable.\bb
\end{abstract}




\maketitle

\section{Introduction}

Lung cancer remains a leading cause of cancer-related morbidity and mortality~\cite{cancerobservatory}. 
Non-small cell lung cancer (NSCLC) accounts for around 85\% of cases, with adenocarcinoma (ADC) and squamous cell carcinoma (SCC) being the most prevalent histological subtypes~\cite{travis2015}. 
Accurate differentiation between ADC and SCC is crucial for clinical decision-making as they are associated with different clinical outcomes~\cite{wang2020comparison} and therapeutic options, including eligibility for specific targeted therapies and immunotherapy strategies~\cite{chansky2009international,campbell2016distinct}. 

Histopathology through a tissue biopsy is the gold standard for distinguishing ADC from SCC, as well as for identifying molecular targets for therapies. Although a CT-guided lung biopsy is highly accurate, it can be associated with complications related to the procedure, such as pneumothorax and bleeding. The risk of technical difficulties and complications increases with small or deep peripheral lesions, challenging biopsy needle trajectories, and in patients with limited cardiopulmonary reserve or an increased risk of bleeding. Additionally, intratumoral heterogeneity can occasionally affect diagnostic accuracy~\cite{de2016image}. These limitations highlight the need for non-invasive, complementary approaches to support subtype assessment when pathology is limited or when additional information would be clinically useful. 

Medical imaging plays a key role in lung cancer. In particular, [$^{18}$F]Fluorodeoxyglucose (FDG) Positron Emission Tomography combined with Computed Tomography (PET/CT) is a standard tool for characterising and staging NSCLC~\cite{takeuchi2014impact}, by combining the metabolic data from PET with the  structural detail of CT~\cite{antoch2003non}. While  ADC and SCC exhibit different metabolic and morphological patterns, discrimination between subtypes based on visual interpretation can sometimes be challenging, resulting in low specificity for subtype classification~\cite{jiang2014thin}. 

Despite its diagnostic advantages, access to PET/CT may vary across institutions due to cost and resource availability. 
Virtual scanning provides a solution to these barriers by using computational models to create synthetic medical images, 
\rr providing a computational estimation of metabolic patterns derived from existing anatomical data\bb
~\cite{wang2022development}.
In this context, virtual scanning is defined as the computational process of mapping structural imaging inputs, such as anatomical CT, to functional imaging outputs, such as synthetic PET volumes.
Such techniques utilize image-to-image translation to synthesize data across modalities, allowing the model to transform source images into a target domain while ensuring that essential diagnostic features remain intact~\cite{isola2017image,rofena2024deep}.
These advancements enable the transformation of anatomical CT data into functional PET representations, providing a non-invasive means to \rr estimate auxiliary metabolic patterns \bb while reducing both scanning costs and patient risk~\cite{dar2019image,guarrasi2025whole}. 
It should be noted, however, that synthetic PET may not serve as a direct replacement for clinical PET in terms of absolute SUV quantification due to the inherent domain shift between generative and biological signals; nonetheless, these synthetic volumes could provide high-value feature representations that benefit downstream diagnostic tasks, such as histological subtyping.

The automated classification of NSCLC subtypes has been extensively explored, primarily through radiomics and deep learning analysis of standalone CT or integrated PET/CT scans. 
Early studies utilized handcrafted radiomic features to identify texture and shape differences between ADC and SCC~\cite{zhu2018radiomic, wu2016exploratory}. 
More recently, convolutional neural networks (CNNs) have demonstrated superior performance by automatically extracting high-dimensional features from CT volumes to predict histology non-invasively~\cite{tomassini2022lung, aksu2025nsclc}. 
Research incorporating metabolic data from PET has further shown that SUV-based metrics and multi-modal fusion can improve classification accuracy over CT alone, as ADC and SCC often exhibit distinct glucose metabolism patterns~\cite{qin2020fine,aksu2024toward}. 
A particularly effective architecture for this task is MINT~\cite{aksu2025multi}, which employs a multi-stage intermediate fusion strategy. 
This framework integrates PET and CT features at varying levels of abstraction, facilitating a gradual and hierarchical synthesis of cross-modality information. 
Such intermediate fusion techniques have proven superior in the biomedical domain, as they effectively exploit the complementary knowledge found in disparate modalities compared to simple early or late fusion methods~\cite{guarrasi2025systematic}.

However, existing literature remains heavily dependent on the availability of real PET/CT data. 
While synthetic PET images have been evaluated for their clinical utility in tasks such as brain disease classification and SUV accuracy~\cite{dayarathna2024deep}, the extension of these generative models to lung cancer histology remains unvalidated. 
Subtype classification represents a more complex downstream task than SUV reconstruction or lesion detection; while the latter focuses on approximating intensity values or localizing hypermetabolic regions, subtyping requires the model to preserve high-dimensional metabolic patterns and intra-tumoral heterogeneity that distinguish ADC from SCC. 
Consequently, it is not yet established whether pseudo-metabolic features possess sufficient discriminatory power to differentiate these subtypes in the absence of real functional imaging.


\rr This study evaluates the utility of synthetic PET features as a supplementary data stream, investigating whether GAN-generated metabolic cues can provide complementary signal to structural CT scans within a multimodal learning framework. \bb
Our primary contributions are as follows:
\begin{itemize}
    \item \rr We investigate the potential of synthetic PET features to improve the discriminatory signal of CT-only models for histological subtyping.\bb
    \item We establish a methodological pipeline that enables the use of multimodal fusion frameworks in clinical settings where physical PET imaging is unavailable.
\end{itemize}

\section{Method}

\begin{figure*}[h]
  \centering
  \includegraphics[width=\textwidth]{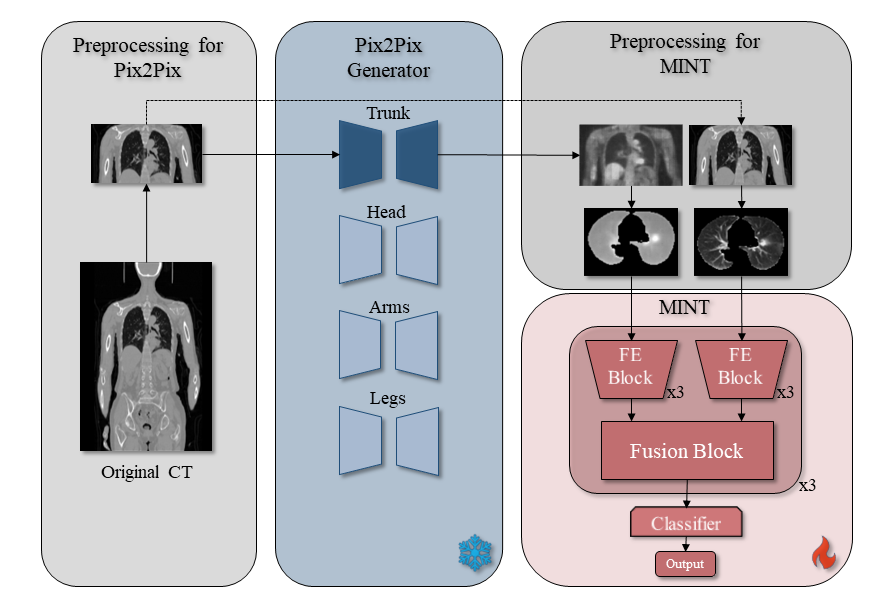}
  \caption{Overview of the proposed framework. The pipeline consists of two primary stages: Synthetic PET generation and Multimodal histological subtyping. FE: Feature Extraction}
  \label{fig:framework}
  \Description{Put a description explaining the graph.}
\end{figure*}

The methodology of this study is organized into five main phases. 
First, we describe the composition of the multi-center dataset and the selection criteria for ADC and SCC cases in Section~\ref{sec:dataset}. 
Second, we detail the preprocessing pipeline used to standardize CT and synthetic PET volumes in Section~\ref{sec:preprocess}. 
Third, we outline the virtual PET synthesis process using a pretrained generative model in Section~\ref{sec:generator}. 
Fourth, we present the architecture of the MINT framework for histological classification in Section~\ref{sec:classifier}. 
Finally, we detail the experimental and training configurations, including the evaluation metrics, in Section~\ref{sec:experiments}.
A general overview of the framework can be found in Figure~\ref{fig:framework}.

\subsection{Dataset}
\label{sec:dataset}

\begin{table}[ht]
\centering
\caption{Summary of the included datasets, showing the total number of patients and the distribution of histological subtypes.}
\label{tab:dataset_summary}
{
\begin{tabular}{lccc}
\hline
\textbf{Dataset} & \textbf{Total Patients} & \textbf{ADC} & \textbf{SQC}  \\
\hline
Humanitas & 423 & 312 (74\%) & 111 (26\%) \\
NSCLC-Radiogenomics & 193 & 160 (83\%) & 33 (17\%)  \\
Lung-PET-CT-Dx & 98 & 74 (76\%) & 24 (24\%) \\
\hline
\end{tabular}
}
\end{table}

The study cohort was compiled by merging an institutional dataset with two public repositories, totaling 714 subjects. 
The distribution of these cases across the different cohorts and histological subtypes is detailed in Table~\ref{tab:dataset_summary}.
A consistent selection protocol was applied across all sources to ensure the integrity of the radiological images and their corresponding histological diagnoses.

The primary subset was obtained from IRCCS Humanitas Research Hospital and initially included 423 patients~\cite{kirienko2018prediction}. 
Eligibility required a diagnosis of NSCLC confirmed by histopathology, a baseline [$^{18}$F]FDG PET/CT scan, and subsequent surgery at the same center. 
Subjects were excluded if they presented with histological types other than ADC or SCC, synchronous malignancies, or a history of cancer within three years of their NSCLC diagnosis. 
These criteria resulted in 312 ADC and 111 SCC cases from this institution.

Additional cases were sourced from the NSCLC Radiogenomics dataset~\cite{radiogenomics}, involving patients from Stanford University and the Palo Alto Veterans Affairs Healthcare System.
Following the established inclusion criteria, 193 patients were selected, consisting of 160 ADC and 33 SCC cases. 
Finally, data from the Lung-PET-CT-Dx repository~\cite{lung-pet-ct-dx} was incorporated; only ADC and SCC cases with paired CT and PET imaging were included, contributing 74 ADC and 24 SCC cases. 
The aggregated dataset used for this study comprised 546 ADC and 168 SCC cases.

\subsection{Preprocessing}
\label{sec:preprocess}

The preprocessing pipeline is divided into two distinct workflows to accommodate the dual-stage architecture of the framework. 
The first stage prepares CT scans for the Pix2Pix generator~\cite{guarrasi2025whole}. Intensity values are standardized by converting CT data to Hounsfield Units (HU), then normalized to a standard intensity range of [0, 1] using min-max scaling. 
To ensure the generative model focuses on relevant pathology, only the cropped lung region is fed to the trunk generator; the peripheral anatomical regions and secondary generators within the Pix2Pix suite are not utilized in this study.

The second stage prepares the data for the MINT classification framework~\cite{aksu2025multi} once the Pix2Pix trunk model generates the synthetic PET volumes. 
This process narrows the focus to the biological region of interest to facilitate more effective feature extraction. 
A well-established segmentation algorithm~\cite{hofmanninger2020automatic} is utilized to extract lung volumes from the CT scans, and the resulting masks are applied to both the CT and the synthetic PET volumes. 
Because the synthetic PET volumes are generated directly from preprocessed CT data, they maintain inherent spatial alignment with the structural input, eliminating the need for additional registration steps before multimodal fusion.

\subsection{Synthetic PET Generation}
\label{sec:generator}

For the metabolic imaging component, we adopted a specialized multi-district generative framework~\cite{guarrasi2025whole}. This framework addresses the anatomical and functional heterogeneity of the human body by decomposing the whole-body image-to-image translation task into region-specific sub-tasks. The architecture first utilizes an automated segmentation stage, where the MOOSE segmentator~\cite{sundar2022fully} partitions a 3D CT volume into four distinct anatomical districts: head, trunk, arms, and legs. For each district, a dedicated 3D Generative Adversarial Network (GAN) based on the Pix2Pix architecture~\cite{isola2017image} is employed to perform the CT-to-PET translation.
In the original development of this framework , each district-specific model was trained on the FDG-PET/CT Lesions dataset~\cite{fdgpetct} using a 3D patch-based approach. The volumes were divided into 32×32×32 voxel patches during training to manage the computational complexity of volumetric data. 

In our implementation, as the focus is restricted to pulmonary oncology, we exclusively utilize the trunk generator to process the thoracic volumes. The process follows the inference pipeline of the original framework, utilizing a sliding window technique with a window size of 32×32×32 and an overlap of 16 voxels. The generator translates these structural CT patches into synthetic metabolic PET patches. These individual patches are then reassembled into a single synthetic volume using a stitching method that averages overlapping regions, ensuring smooth spatial transitions and the elimination of edge artifacts. This allows for the reconstruction of a continuous synthetic PET volume that maintains anatomical correctness and spatial alignment with the original CT scan.
\rr In our implementation, the trunk generator was intentionally kept frozen to strictly evaluate the generalizability of the learned structural-to-functional mapping across heterogeneous clinical environments. This decision serves as a methodological safeguard to prevent potential data leakage; by avoiding center-specific fine-tuning on the target multi-center cohort, we ensure that the MINT framework extracts diagnostic features from underlying biological patterns rather than GAN-specific artifacts or center-dependent generative signatures. \bb

\subsection{Histological Subtype Classification}
\label{sec:classifier}

For the classification task, we employed the MINT framework, a multi-stage intermediate fusion architecture specifically developed for the differentiation of NSCLC histological subtypes~\cite{aksu2025multi}. 
The model architecture is organized into sequential stages, each containing independent feature extraction blocks dedicated to processing each modality. 
Following the extraction of modality-specific features, a fusion layer integrates the representations through element-wise multiplication. 
These fused features are then reintegrated into the original modality streams via element-wise addition, acting as a residual-like mechanism across stages. 
This hierarchical process is repeated throughout the network to facilitate gradual cross-modal integration. 
In accordance with the findings of the original study, we utilized a three-stage configuration with three sets of feature extraction blocks per stream, as this architecture was shown to achieve optimal performance in subtype classification.

The fundamental units for feature representation in the MINT framework are modeled after the 3D ResNet "basic block" architecture~\cite{tran2018closer}. 
This structure employs a dual-branch design, comprising a primary convolutional path and a residual branch, which facilitates the training of deeper network hierarchies by addressing the vanishing gradient problem.
Within each block, the primary branch performs sequential operations: an initial $3 \times 3 \times 3$ convolution, batch normalization, and ReLU activation, followed by a second $3 \times 3 \times 3$ convolutional layer and batch normalization. 
To maintain spatial and channel-wise alignment, the initial block of each stage utilizes a $1 \times 1 \times 1$ convolution within the residual branch. 
This alignment is critical because the first convolutional layer in each stage applies a stride of 2, effectively halving the spatial dimensions while doubling the feature map depth. 
This approach serves as an alternative to explicit pooling layers, ensuring that the network maintains high representational capacity as it transitions through the hierarchy.
The outputs from the two branches are integrated via element-wise summation before a final ReLU activation is applied. 
In the subsequent blocks within a stage, a stride of 1 is maintained to preserve spatial dimensions, allowing the parallel branch to function as a standard skip connection. 
This modular design allows the output of any given block to be passed either to the succeeding extraction unit or directly to a fusion module, depending on its position within the multi-stage hierarchy.

The integration of CT and PET data is performed within specialized fusion blocks designed to exploit the complementary nature of anatomical and functional information. 
These blocks facilitate a bidirectional exchange of features between the unimodal streams. 
To generate a unified cross-modal representation, the feature maps from the final extraction block of each stage $s \in \{1, 2, 3\}$, denoted as $\mathbf{X}_{CT}^{(s)}$ and $\mathbf{X}_{PET}^{(s)}$, are first processed through a $1 \times 1 \times 1$ convolutional layer. 
This operation functions as a channel-wise bottleneck, reducing the depth of each modality's representation to a single feature map.
Following batch normalization ($BN$), the two down-sampled feature maps are integrated using element-wise multiplication ($\odot$) to produce a joint cross-modal descriptor, $\mathbf{Z}^{(s)}$. 
This multiplicative interaction highlights regions of high importance shared across both modalities. 
This fused representation is then reintegrated into the original unimodal branches through element-wise summation ($\oplus$), effectively acting as a residual connection that enriches the modality-specific features with cross-modal context. 
The process is defined by the following equations:

\begin{displaymath}
    \mathbf{Z}^{(s)} = BN(\omega(\mathbf{X}_{CT}^{(s)})) \odot BN(\omega(\mathbf{X}_{PET}^{(s)}))
\end{displaymath}

\begin{displaymath}
    \mathbf{\hat{X}}_{CT}^{(s)} = \mathbf{X}_{CT}^{(s)} \oplus \mathbf{Z}^{(s)}
\end{displaymath}

\begin{displaymath}
    \mathbf{\hat{X}}_{PET}^{(s)} = \mathbf{X}_{PET}^{(s)} \oplus \mathbf{Z}^{(s)}
\end{displaymath}

where $\omega$ represents the $1 \times 1 \times 1$ convolutional operation. 
The resulting maps, $\mathbf{\hat{X}}_{CT}^{(s)}$ and $\mathbf{\hat{X}}_{PET}^{(s)}$, carry integrated information forward to the subsequent stage of the hierarchy, allowing for a progressive refinement of the multi-modal features.

Following the final fusion stage ($s=3$), the refined feature maps from both the CT and PET streams, $\mathbf{\hat{X}}_{CT}^{(3)}$ and $\mathbf{\hat{X}}_{PET}^{(3)}$, are concatenated along the channel dimension. 
To transition from high-dimensional feature maps to a compact representation suitable for classification, a 3D global average pooling layer is applied. 
This operation computes the spatial average of each feature map, resulting in a single feature vector that summarizes the global characteristics of the lung volumes across both modalities.
The resulting aggregated vector is passed through a fully connected layer, which maps the integrated features to the two target classes: ADC and SCC. 
Finally, a Softmax activation function is applied to produce the class-wise probability scores, enabling the non-invasive prediction of the histological subtype. 
The entire network is trained end-to-end to minimize the cross-entropy loss between the predicted probabilities and the ground-truth pathological labels.

\subsection{Experimental Configuration}
\label{sec:experiments}

For a direct comparison across unimodal (CT-only) and multimodal (CT and PET) configurations, we employed an identical stratified 5-fold cross-validation scheme at the patient level.
The aggregated dataset was shuffled and partitioned into training, validation, and test sets with a 60/20/20 ratio, respectively. 
This stratification ensured that the class proportions of ADC and SCC remained consistent across all folds.

Following the optimization results reported in the baseline MINT study~\cite{aksu2025multi}, we employed a three-stage architecture with three feature extraction blocks per stage. 
The network depth was configured with 16 initial feature maps in the first stage, which were doubled at each subsequent stage, resulting in 32 maps for the second stage and 64 for the third. 
The final multimodal feature vector, obtained after concatenating the global average pooled outputs of both streams, comprised 128 features.
Training was conducted over 100 epochs using the Adam optimizer with an initial learning rate of 0.001. 
To facilitate convergence, the learning rate was decayed by a factor of 0.1 every 25 epochs. 
Given the inherent class imbalance between ADC and SCC cases, class weights were incorporated into the cross-entropy loss function.

Model performance was assessed using the following metrics: the Area Under the Receiver Operating Characteristic Curve (AUC), Balanced Accuracy, the Geometric Mean (Gmean), and the True Positive Rate (TPR) for both ADC and SCC subtypes. 
While AUC evaluates the model’s discriminative ranking capability regardless of class distribution, the remaining metrics were selected to provide a balanced assessment of performance given the inherent class imbalance in our dataset. 
These metrics are defined based on True Positives (TP), True Negatives (TN), False Positives (FP), and False Negatives (FN):

\begin{displaymath}
    \text{Balanced Accuracy} = \frac{1}{2} \left( \frac{TP}{TP + FN} + \frac{TN}{TN + FP} \right)
\end{displaymath}

\begin{displaymath}
    \text{GMean} = \sqrt{\frac{TP}{TP + FN} \times \frac{TN}{TN + FP}}
\end{displaymath}

\begin{displaymath}
    \text{TPR} = \frac{TP}{TP + FN}
\end{displaymath}

Additionally, to validate the fidelity of the generated pseudo-PET volumes against the available ground-truth PET scans, we utilized standard image synthesis 
metrics~\cite{di2023comparative}. 
Mean Absolute Error (MAE) measures the average voxel-wise intensity difference to assess quantitative accuracy. 
Peak Signal-to-Noise Ratio (PSNR) evaluates the ratio between the maximum possible signal power and the power of corrupting noise, while the Structural Similarity Index (SSIM) assesses the preservation of luminance, contrast, and structural information. 
These metrics are defined as follows:

\begin{displaymath}
    \text{MAE} = \frac{1}{N} \sum_{i=1}^{N} |y_i - \hat{y}_i|
\end{displaymath}

\begin{displaymath}
    \text{PSNR} = 10 \cdot \log{10} \left( \frac{MAX_I^2}{MSE} \right
)\end{displaymath}

\begin{displaymath}
    \text{SSIM}(x, y) = \frac{(2\mu_x\mu_y + c_1)(2\sigma_{xy} + c_2)}{(\mu_x^2 + \mu_y^2 + c_1)(\sigma_x^2 + \sigma_y^2 + c_2)}
\end{displaymath}
In these definitions, $y$ and $\hat{y}$ represent the ground-truth and synthetic PET voxel intensities, respectively, while $N$ denotes the total number of voxels. 
For the SSIM calculation, $\mu$ and $\sigma$ represent the local means and standard deviations, while $c_1$ and $c_2$ are constants included to ensure numerical stability.
For PSNR, $MAX_I^2$ denotes the maximum possible voxel intensity after normalization.

\section{Results}

In this section, we present the experimental findings of our proposed framework for NSCLC histological subtyping. 
We begin with a two-part evaluation of the synthesized modality in Section~\ref{sec:res_pet}, assessing both the quantitative fidelity and the qualitative biological plausibility of the generated volumes. 
Following this, Section~\ref{sec:res_classification} details the performance of the MINT framework in the classification of ADC and SCC, highlighting the performance gains achieved through multimodal fusion. 

\subsection{Evaluation of Synthetic PET Quality}
\label{sec:res_pet}

\begin{table*}
  \caption{Quantitative performance of the synthetic PET generation pipeline across different datasets. Metrics were calculated for each individual scan, with the table reporting the mean $\pm$ standard deviation. Note that the FDG-PET/CT Lesions metrics are cited from the original publication~\cite{guarrasi2025whole} for comparative context.}
  \label{tab:res_pet}
  \begin{tabular}{cccc}
    \toprule
    Dataset & PSNR $\uparrow$ & SSIM $\uparrow$ & MAE $\downarrow$ \\
    \midrule
    Humanitas & 37.435 $\pm$ 2.201 & 0.848 $\pm$ 0.052 & 0.007 $\pm$ 0.010 \\
    NSCLC Radiogenomics & 34.837 $\pm$ 1.441 & 0.805 $\pm$ 0.034 & 0.009 $\pm$ 0.002 \\
    Lung-PET-CT-Dx & 34.445 $\pm$ 1.280 & 0.835 $\pm$ 0.047 & 0.007 $\pm$ 0.001  \\
    Combined & 36.306 $\pm$ 2.339 & 0.835 $\pm$ 0.047 & 0.007 $\pm$ 0.007 \\
    \midrule
    FDG-PET/CT Lesions & 32.360 $\pm$ 1.780 & 0.870 $\pm$ 0.020 & 0.008 $\pm$ 0.001\\
  \bottomrule
\end{tabular}
\end{table*}

The quantitative fidelity of the synthesized PET volumes was assessed by calculating three standard image quality metrics for each individual scan: PSNR, SSIM, and MAE. 
Table~\ref{tab:res_pet} summarizes these results, reporting the average and standard deviation across all cases in each respective dataset. 
These metrics provide a voxel-wise comparison between the generated pseudo-PET volumes and the clinical ground-truth PET scans.
As shown in Table~\ref{tab:res_pet}, the Pix2Pix trunk generator demonstrated high performance in reconstructing metabolic distributions from structural CT data, achieving an aggregate PSNR of $36.306 \pm 2.339$ dB and a consistently low MAE of $0.007 \pm 0.007$. 
The highest fidelity was observed in the Humanitas dataset, which yielded the highest average PSNR ($37.435 \pm 2.201$) and SSIM ($0.848 \pm 0.052$). 
Furthermore, the performance across the Humanitas, NSCLC Radiogenomics, and Lung-PET-CT-Dx datasets outperformed the original benchmark metrics reported for the FDG-PET/CT Lesions dataset~\cite{guarrasi2025whole} in terms of PSNR, suggesting that the frozen trunk generator generalizes effectively to targeted thoracic volumes.
The consistently low MAE values (ranging from 0.007 to 0.009) and high PSNR values across all cohorts indicate that the synthetic volumes provide a reliable voxel-wise approximation of tracer uptake.
While a slight decrease in PSNR was noted in the Lung-PET-CT-Dx dataset ($34.445 \pm 1.280$), the structural similarity remained high ($0.835 \pm 0.047$), confirming that the model maintains stable performance across different patient cases and acquisition protocols. 
These results establish a robust foundation for the subsequent feature extraction and histological classification stages of the framework.

\begin{figure}[h]
  \centering
  \includegraphics[width=\linewidth]{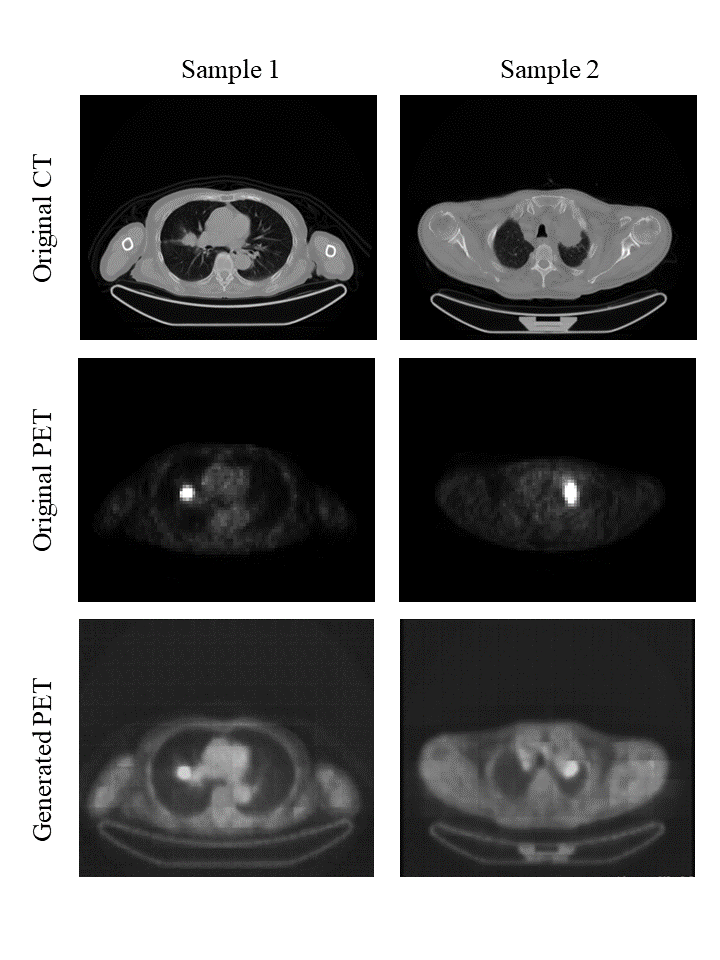}
  \caption{Representative axial slices from two distinct cases illustrating the qualitative performance of the PET synthesis pipeline. From top to bottom: Original structural CT input; Clinical [$^{18}$F]FDG-PET ground truth; and Synthesized PET output.}
  \label{fig:sample_images}
  \Description{Put a description explaining the graph.}
\end{figure}

While the quantitative metrics establish a statistical baseline for the synthesis performance, a qualitative visual inspection is essential to confirm the biological plausibility of the generated volumes. 
As illustrated in Figure~\ref{fig:sample_images}, the 3D Pix2Pix generator demonstrates a sophisticated ability to map structural CT features to functional pseudo-metabolic representations. 
Specifically, in cases of confirmed NSCLC, the synthetic PET volumes consistently exhibit high-intensity focal points in the tumor regions, accurately reflecting the expected hypermetabolic state associated with malignant growth.

Beyond the primary lesion, the model successfully approximates the physiological distribution of the radiopharmaceutical, showing realistic low-level tracer uptake in healthy lung parenchyma and appropriate structural contrast. 
Visual assessment confirmed that the synthetic modality preserves the spatial morphology of the tumor, which is critical for the MINT framework's ability to extract discriminatory features. 
\rr The alignment of localized intensities with the tumor site suggests the model has internalized the relationship between tissue morphology and uptake, providing an auxiliary metabolic representation that effectively supplements structural CT data for multimodal classification.\bb

\subsection{Histological Classification Performance}
\label{sec:res_classification}

\begin{table*}[t]
  \caption{Performance comparison of histological subtype classification (ADC vs. SCC) using the multimodal approach (CT + Synthetic PET) versus CT-only. Values represent the mean $\pm$ standard deviation across cross-validation folds. Bold values in the metric rows indicate superior performance, while bold p-values denote statistical significance ($p < 0.05$) based on the Wilcoxon signed-rank test. AUC: Area under the receiver operating characteristic curve; GMean: Geometric mean of true positive rates of ADC and SCC; ADC: Adenocarcinoma; SCC: Squamous Cell Carcinoma; TPR: True Positive Rate.}
  \label{tab:res_hst}
  \begin{tabular}{cccccc}
    \toprule
    Model & AUC & Balanced Accuracy & GMean & TPR for ADC & TPR for SCC \\
    \midrule
    Multimodal & \textbf{0.591 $\pm$ 0.021} & \textbf{0.555 $\pm$ 0.037} & \textbf{0.524 $\pm$ 0.065} & 0.599 $\pm$ 0.135 & \textbf{0.511 $\pm$ 0.201}  \\
    CT-only & 0.489 $\pm$ 0.086 & 0.516 $\pm$ 0.034 & 0.305 $\pm$ 0.253 & \textbf{0.690 $\pm$ 0.283} & 0.343 $\pm$ 0.289 \\
    \midrule
    p-value & \textbf{0.03125} & 0.06250 & \textbf{0.03125} & 0.84375 & 0.21875 \\

  \bottomrule
\end{tabular}
\end{table*}

We evaluated the impact of incorporating synthetic PET data on the classification of ADC and SCC subtypes. 
The goal was to determine if the addition of a synthesized metabolic modality provides the MINT framework with sufficient complementary information to outperform the CT-only baseline.

The results in Table \ref{tab:res_hst} demonstrate that the integration of synthetic PET data enhances the model's ability to distinguish between NSCLC histological subtypes compared to using CT data alone. 
While the CT-only baseline achieved an AUC of $0.489 \pm 0.086$, essentially performing at the level of random chance, the Multimodal approach reached an AUC of $0.591 \pm 0.021$, a statistically significant improvement ($p = 0.03125$).
\rr The near-random performance of the CT-only baseline ($AUC=0.489$) underscores the inherent difficulty of histological subtyping from anatomical data alone and highlights the necessity of the proposed virtual scanning augmentation to extract otherwise inaccessible discriminatory cues. \bb

The impact of the synthetic modality is specifically evident in the improved balance between class-specific metrics. 
The CT-only model exhibited a bias toward ADC, with a TPR of $0.690$ but a lower TPR of $0.343$ for SCC. 
The Multimodal approach mitigated this imbalance by increasing the SCC detection rate to $0.511$, representing a nearly $50\%$ improvement in sensitivity for that class. 
This change is further reflected in the GMean, which increased from $0.305$ to $0.524$ ($p = 0.03125$). 
These findings suggest that the synthetic PET generated via Pix2Pix provides a useful second modality that enables the MINT framework to utilize its multimodal fusion capabilities, whereas CT alone provides insufficient information for stable classification.

\rr To further isolate the contribution of the metabolic modality, we evaluated the classification performance using PET data exclusively. As summarized in Table~\ref{tab:real_fake_pet}, neither the Real PET nor the Synthetic PET models were able to achieve reliable discrimination between ADC and SCC when used in a unimodal configuration. The Real PET baseline yielded an AUC of $0.465 \pm 0.137$, while the Synthetic PET reached $0.459 \pm 0.031$, both performing slightly below the threshold of random chance. When compared to the CT-only baseline (Table~\ref{tab:res_hst}), which also struggled to maintain stability ($AUC = 0.489$), it becomes evident that no single modality contains sufficient diagnostic features for this task. These results underscore the necessity of a multimodal framework; it is not the synthetic modality itself that drives performance, but rather the synergistic integration of structural cues from CT and the pseudo-metabolic features from synthetic PET that enables the MINT framework to surpass the limitations of individual imaging domains.

\begin{table}[h]
  \caption{\rr Classification performance using PET-only modalities. Values represent the mean $\pm$ standard deviation. AUC: Area under the receiver operating characteristic curve; BACC: Balanced accuracy; GMean: Geometric mean of true positive rates of ADC and SCC. \bb}
  \label{tab:real_fake_pet}
  \begin{tabular}{cccc}
    \toprule
    Model & AUC & BACC & GMean \\
    \midrule
    Real PET & \textbf{0.465 $\pm$ 0.137} & 0.478 $\pm$ 0.085 & \textbf{0.329 $\pm$ 0.098} \\
    Synthetic PET & 0.459 $\pm$ 0.031 & \textbf{0.499 $\pm$ 0.010} & 0.164 $\pm$ 0.145  \\
  \bottomrule 
\end{tabular}
\end{table}

We further investigated the architectural requirements for effective multimodal integration by comparing three distinct fusion strategies: Early, Late, and Intermediate (MINT) fusion. As shown in Table~\ref{tab:fusions}, the choice of fusion mechanism significantly impacts the model's ability to leverage the synthetic PET modality.
Late fusion, which aggregates independent predictions from each modality, failed to achieve stable results ($AUC = 0.458 \pm 0.104$, $GMean = 0.076 \pm 0.151$), confirming that neither CT nor Synthetic PET contains sufficient independent signal for high-level decision-level fusion.
While Early fusion (input-level concatenation) showed competitive performance ($AUC = 0.581$), it exhibited higher variance across cross-validation folds. 
The multi-stage intermediate fusion approach utilized by the MINT framework achieved the highest performance across all metrics ($AUC = 0.591$, $GMean = 0.524$) with the lowest standard deviation. 
These results indicate that hierarchical integration across multiple abstraction levels is superior to simple input or decision-level fusion. 
By performing fusion at sequential stages of the feature extraction process, the model is able to capture complex, non-linear correlations between structural CT features and synthesized metabolic signals at varying spatial scales and semantic depths. 
This gradual synthesis preserves cross-modal information that is otherwise lost during the rigid input-level concatenation of Early fusion or the independent prediction-averaging of Late fusion.

\begin{table}[h]
  \caption{\rr Impact of fusion methods on classification performance. Values represent the mean $\pm$ standard deviation. AUC: Area under the receiver operating characteristic curve; BACC: Balanced accuracy; GMean: Geometric mean of true positive rates of ADC and SCC. \bb}
  \label{tab:fusions}
  \begin{tabular}{cccc}
    \toprule
    Model & AUC & BACC & GMean \\
    \midrule
    Early & 0.581 $\pm$ 0.075 & 0.554 $\pm$ 0.057 & 0.502 $\pm$ 0.075 \\
    Late & 0.458 $\pm$ 0.104 & 0.489 $\pm$ 0.023 & 0.076 $\pm$ 0.151 \\
    MINT &  \textbf{0.591 $\pm$ 0.021} & \textbf{0.555 $\pm$ 0.037} & \textbf{0.524 $\pm$ 0.065} \\
  \bottomrule 
\end{tabular}
\end{table}

\begin{figure}[h]
  \centering
  \includegraphics[width=\linewidth]{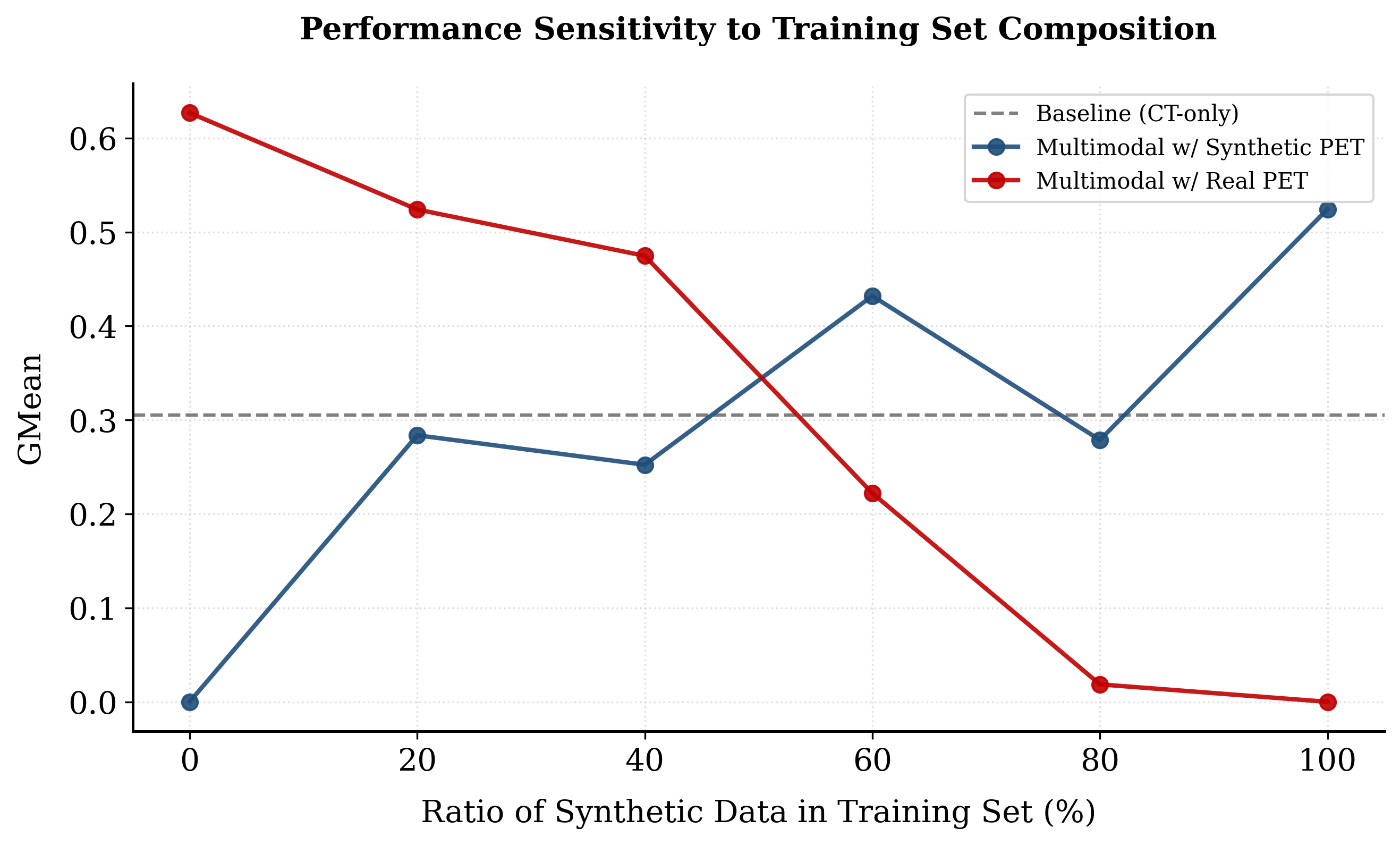}
  \caption{Impact of training set composition on GMean performance across real and synthetic testing domains. The dashed line indicates the CT-only baseline. The red line represents performance on real PET test data, while the blue line represents performance on synthetic PET test data.}
  \label{fig:ratios_gmean}
  \Description{Put a description explaining the graph.}
\end{figure}

\begin{figure}[h]
  \centering
  \includegraphics[width=\linewidth]{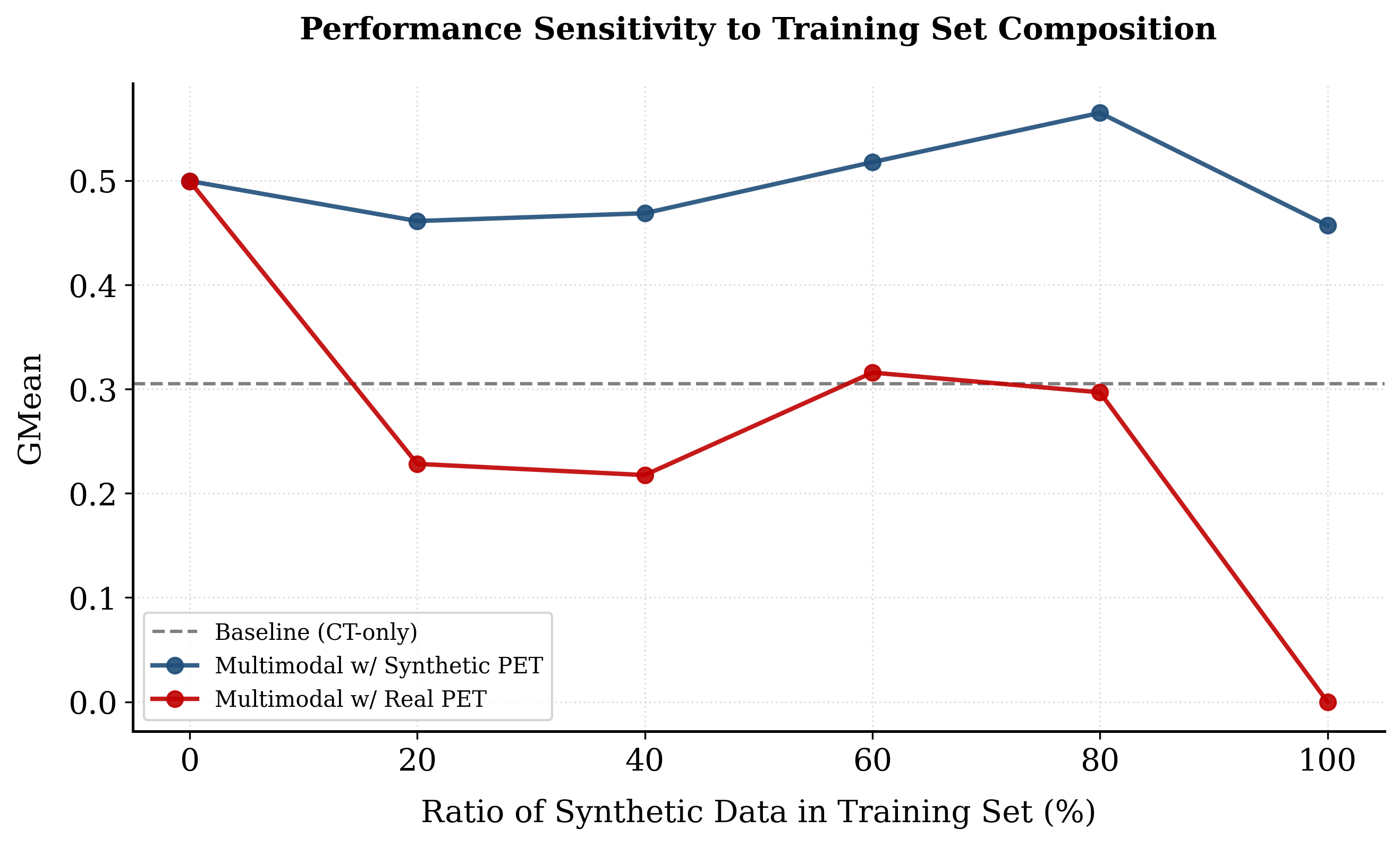}
  \caption{Sensitivity analysis of GMean performance utilizing CLAHE-enhanced synthetic volumes.}
  \label{fig:ratios_gmean_clahe}
  \Description{Put a description explaining the graph.}
\end{figure}

\bb To evaluate the representational consistency of the synthesized modality and its compatibility with biological imaging, we conducted an exploratory sensitivity analysis by systematically varying the composition of the training dataset. 
As shown in the Figures~\ref{fig:ratios_gmean} and~\ref{fig:ratios_gmean_clahe}, these additional experiments utilized mixed training sets containing both real and synthetic PET volumes, with the ratio of synthetic data systematically increased from 0 to 100 along the x-axis. 
This analysis sought to determine if the pseudo-metabolic features learned by the 3D Pix2Pix generator are interchangeable with biological uptake signals or if they represent a distinct generative domain.

The results of this analysis reveal a pronounced dependency on the training domain. 
When evaluated on real PET test data, performance peaked at a 0 ratio, corresponding to an entirely real training set. 
However, as the proportion of synthetic data in the training set increased, performance on real scans steadily declined, reaching near-zero values at the 100 ratio. 
This trend suggests a significant domain shift, indicating that the synthetic features learned by the model do not perfectly mirror the absolute characteristics found in physical clinical PET imaging.

Conversely, performance on synthetic PET testing remained unstable or below the structural baseline when the training set was dominated by real scans. 
A significant performance gain was observed as the ratio approached 100, where the Gmean reached its peak of 0.524. 
The dashed horizontal line in the figures represents the performance of the unimodal CT-only branch, which maintains a Gmean of 0.305. 
In particular, the multimodal model trained exclusively on synthetic data substantially outperformed this structural baseline when evaluated in the virtual scanning domain. 
\rr While the performance decline on real PET scans confirms a significant domain shift, the statistically significant gain in multimodal GMean (Table 3) suggests the GAN effectively captures 'discriminatory metabolic proxies.' These generative patterns, while not identical to absolute SUV values, provide biologically grounded signals that structural CT lacks. \bb

In a further exploration of these domain dynamics, we observed that the application of Contrast Limited Adaptive Histogram Equalization (CLAHE) appeared to influence performance stability.
As seen in Figure~\ref{fig:ratios_gmean_clahe}, incorporating CLAHE in both the training and testing phases for synthetic data tended to smooth the performance fluctuations observed in the vanilla results.
While the primary findings of this study rely on the standardized 100\% synthetic pipeline, these observations suggest that targeted histogram processing can partially mitigate the impact of the domain shift by standardizing the feature distributions across different ratios.

\rr
Finally, to contextualize the performance of the proposed virtual scanning pipeline, we compared the results of the multimodal framework using both real and synthetic PET data. As shown in Table~\ref{tab:original}, the integration of structural CT with biological [$^{18}$F]FDG signals (CT + Real PET) achieves the highest classification performance ($AUC = 0.681 \pm 0.042$, $GMean = 0.646 \pm 0.062$), underscoring the diagnostic superiority of physical metabolic imaging. However, while the synthetic modality (CT + Synt. PET) does not fully surrogate the predictive power of real PET, it maintains a statistically significant lead over the CT-only baseline ($AUC = 0.591$ vs $0.489$). These results suggest that in clinical environments where the acquisition of real PET/CT is restricted by high costs, logistical barriers, or patient contraindications, synthetic PET provides a crucial auxiliary data stream. By effectively bridging the gap between unimodal anatomical scans and full multimodal imaging, virtual scanning enables the extraction of otherwise inaccessible metabolic cues, offering a pragmatic enhancement for AI-assisted oncology workflows where real functional data is unavailable. \bb

\begin{table}[h]
  \caption{\rr Comparison of multimodal results achieved with real and synthetic PETs. Values represent the mean $\pm$ standard deviation. AUC: Area under the receiver operating characteristic curve; BACC: Balanced accuracy; GMean: Geometric mean of true positive rates of ADC and SCC. \bb}
  \label{tab:original}
  \begin{tabular}{cccc}
    \toprule
    Model & AUC & BACC & GMean \\
    \midrule
    CT + Real PET & \textbf{0.681 $\pm$ 0.042} & \textbf{0.661 $\pm$ 0.039} & \textbf{0.646 $\pm$ 0.062} \\
    CT + Synt. PET &  0.591 $\pm$ 0.021 & 0.555 $\pm$ 0.037 & 0.524 $\pm$ 0.065 \\
  \bottomrule 
\end{tabular}
\end{table}

\section{Conclusion}

This study demonstrates that utilizing synthetic PET data as a supplementary modality significantly improves the histological classification of NSCLC when compared to using CT imaging alone. 
By integrating synthetic images generated by a Pix2Pix-based synthesis model with the MINT framework, we achieved a more balanced and accurate differentiation between ADC and SCC subtypes. 
Specifically, the inclusion of synthetic metabolic cues led to a statistically significant increase in AUC ($0.591$) and GMean ($0.524$), effectively mitigating the class bias observed in CT-only models. 

Our findings suggest that while CT scans provide essential anatomical detail, they often lack the discriminatory features necessary for stable histological subtype classification.
Synthetic PET scans provide useful pseudo-metabolic features that allow multimodal architectures like MINT to leverage their full fusion potential and provide better classification performance. 
\rr This study serves as a feasibility assessment for auxiliary feature enhancement. While synthetic PET does not replace the gold standard of biological imaging, it provides a statistically significant lead over CT-only diagnostics. \bb

Despite these promising results, certain limitations inherent to the current generative approach must be noted. 
\rr Our exploratory analysis revealed a domain shift between synthetic and real PET data, suggesting that pseudo-metabolic features do not yet perfectly replicate the absolute Standardized Uptake Values (SUVs) found in clinical scans. \bb
While the application of Contrast Limited Adaptive Histogram Equalization (CLAHE) served to partially mitigate this shift and stabilize performance, the disparity remains a challenge for direct cross-domain transferability. 
Furthermore, the generative model was trained on a dataset predominantly from a single institution, which may affect generalizability across different clinical settings.

Future work will focus on addressing these challenges through several avenues. 
To further reduce the domain gap, we plan to evaluate more advanced generative architectures beyond standard GANs and integrate larger, multi-centric datasets to enhance model robustness. 
Additionally, refining the segmentation process with organ-specific subdivisions may allow for even greater specialization of the synthetic translation. 
These advancements will further the development of healthcare Digital Twins, providing accurate virtual imaging representations to monitor and optimize personalized health outcomes without the need for additional invasive procedures.

\begin{acks}
This work was partially supported by: i) PNRR-MCNT2-2023-12377755, ii) Kempe project JCSMK24-0094. 
Resources are provided by the Swedish National Academic Infrastructure for Supercomputing and by National Infrastructure for Computing (SNIC) at Alvis @ C3SE, funded by the Swedish Research Council through grant agreements no. 2022-06725 and no. 2018-05973.
\end{acks}

\bibliographystyle{ACM-Reference-Format}
\bibliography{bibliography}

\end{document}